# An Integrated Classification Model for Financial Data Mining

**Abstract.** Nowadays, financial data analysis is becoming increasingly important in the business market. As companies collect more and more data from daily operations, they expect to extract useful knowledge from existing collected data to help make reasonable decisions for new customer requests, e.g. user credit category, churn analysis, real estate analysis, etc. Financial institutes have applied different data mining techniques to enhance their business performance. However, simple approach of these techniques could raise a performance issue. Besides, there are very few general models for both understanding and forecasting different financial fields. We present in this paper a new classification model for analyzing financial data. We also evaluate this model with different real-world data to show its performance.

**Keywords:** Data mining, decision tree, multilayer perceptron, Gaussian Process, classification model

## 1 Introduction

Today, we have a deluge of financial datasets. Due to the large sizes of the data sources it is not possible for a human analyst to come up with interesting information (or patterns) that can be used in the decision making process. Global competitions, dynamic markets, and rapid development in the information and communication technologies are some of the major challenges in today's financial industry. For instance, financial institutions are in constant needs for more data analysis, which is becoming more very large and complex. As the amount of data available is constantly increasing, our ability to process it becomes more and more difficult. Efficient discovery of useful knowledge from these datasets is therefore becoming a challenge and a massive economic need.

On the other hand, data mining (DM) is the process of extracting useful, often previously unknown information, so-called knowledge, from large data sets. This mined knowledge can be used for various applications such as market analysis, fraud detection [1], churn analysis [2], etc. DM has also proven to be very effective and profitable in analyzing financial datasets [3]. However, mining financial data presents special challenges; complexity, external factors, confidentiality, heterogeneity, and size. The data miners' challenge is to find the trends quickly while they are valid, as well as to recognize the time when the trends are no longer valid. Besides, designing an appropriate process for discovering valuable knowledge in financial data is also a complex task.

Different DM techniques have been proposed in the literature for data analyzing in various financial applications. For instance, decision-tree [4] and first-order learning [5] are used in stock selection. Neural networks [6] and support vector machine (SVM) [7] techniques were used to predict bankruptcy, nearest-neighbors classification [8] for the fraud detection. Users also have used these techniques for analyzing financial time series [9], imputed financial data [10], outlier detection [11], etc. As different businesses have different behavior-response mapping relationships, and to find a universal fitting model for every particular field is time-consuming if not be impossible, a common approach for mining financial data classification capable of adapting to different business area is needed.

Indeed, as financial dataset is always very large and building a universal model for classification is usually impracticable. A lot of hybrid [12][13] and parallel models [14] for particular financial dataset are developed. However they are not a common structure and do not follow the financial dataset feature, e.g. categorical attributes are summarized concepts, whose rules are uncertain in classification task. On the contrary, numerical data is usually from ETL (**E**xtract, **T**ransform and **L**oad) process and so they are unified in one field. We thus need an approach to minimize using nominal attribute logically and seeking the optimal model for classification to help business instant decision making, e.g. credit risk analysis, customer churn prediction, and house price rank instant notification, etc.

In this paper, we propose a new hybrid classification process that can not only understand and forecast the financial datasets, but also gain useful structural knowledge, e.g. significant nominal groups, tightness of groups. We also evaluate our model with real-world datasets. Indeed, we present the capacity of our model for parallel computing paradigm to speed up the training and analyzing process.

The rest of this paper is organized as follows. In Section 2, we present the technique background of this paper and related work. Next, in Section 3 we describe the combined model for financial dataset classification. We show criteria for partitioning and evaluation scheme for the designed model in Section 4. Experiments and analysis are given in Section 5. Finally, we summarize our work and cite some future work in the last section.

## 2 Background

In this section, we evaluate traditional data mining techniques widely used in analyzing financial datasets: decision tree (DT), Bayes network (BN), clustering, neural networks (NN) and Gaussian process (GP). Next, we resume related work in this context.

### 2.1 Data Mining Techniques

**Decision Tree.** C4.5[15] is a popular decision tree model which greedily chooses largest information gain ratio to build. However, small perturbations in the dataset will probably cause a considerable difference in the produced decision tree. Pruning is used to avoid overfitting but there's no theoretically guarantee about the efficiency. Moreover, 2-way growth of numerical attribute is inefficient.

**Bayes Network** [16] is a probabilistic graphical model representing conditional dependencies via a directed acyclic graph, efficient in learning and can be paralleled easily. However, there is no universally accepted best training method and it involves expert to decide explainable causal influences.

**Clustering**. In the business, clustering can be used, for instance, to segment customers into a number of groups for additional analysis and marketing strategy. Clustering also has its drawbacks, e.g. traditional clustering, as K-means clustering[17], can only handle numerical attributes, and is weak at computing accurate behavior-response mapping relationship since training is unsupervised and dropping targets.

**Neural Networks.** Multi-layer perceptron (MLP) can handle complex classification problem [18][19]. However the cons of MLP are clear: No prior idea of the optimal size of hidden layer. Too small setting will produce very poor network with potential over-generalizing. While too large setting will cause very slow training and many hypeplane may actually coincide after training, and if not, it has over-fitting problem.

**Gaussian Process.** In contrasts with DT and MLP giving only a guess at the class label, Gaussian Process [20] gives the class probabilities as output. Gaussian Process Classification (GPC) equals to 1 hidden layer MLP with infinite number of hidden neurons.

Tractable exact inference is not feasible, latent function with posterior GP can be obtained by Laplace approximate inference with second order Taylor expansion and Newton's method. Possibilities for two classes in binary classification are symmetrical. In this paper, discriminative approach is used to model the target possibility function where $C_1$ stands for class *1*, in contrast with class *-1*, *f* is the latent function.

$$p(C_1 | f(x) = \sigma(f(x)) = \frac{1}{1+\exp(-f(x))} \quad (1)$$

The cons of GPC is its complexity $O(n^3)$, which limits the method when data size is large since both hardware resource and time consuming will increase dramatically.

## 2.2 Related Work

As financial dataset is always very large and building a universal model for classification is usually impracticable and not accurate. A lot of hybrid models for financial dataset are developed, e.g. hybrid model includes, [12] uses rule learners, decision lists, decision tree and association rules. However it mainly replies on nominal labels; [21] uses decision tree and genetic algorithm based hybrid model. But it can only handle small disjunct with a small number of training examples. [13] mixes genetic algorithms with SVM to optimize feature subset and parameters of SVM. However SVMs can only handle numerical attributes and binominal labels; [14] Integrates financial ratios, intellectual capital ratios and neural network. However it only involves numerical ratios and it is not a common structure as [13]. More importantly, they do not follow the financial dataset features, e.g. categorical attributes are summarized concepts, whose rules are uncertain in classification task and can be missing due to record neglect or different operational procedures between branches/companies/periods. On the contrary, numerical data is usually from ETL devices and so they are unified in one field. We thus need a common approach to minimize using nominal attribute logically and seeking the optimal model for classification to help business instant decision making, e.g. credit risk analysis, customer churn prediction, and house price rank instant notification, etc.

## 3 A Combined Model for Financial Dataset Classification

In this section, we present an application of data mining techniques for structural understanding and forecasting financial dataset, which has differently scaled attributes and, consists of both nominal and numerical attributes, assuming similar behavior-response clusters exist. The training and forecast processes are shown in Fig 1.

We derive our scheme as G-KM-NC. The model consists of three parts: G stands for grouping; KM stands for K-means clustering [17] for a particular group. NC stands for non-linear classifier technique, e.g. MLP and GPC, in which KM can be omitted if group is tight by vision or clustering criteria discussed in Section 4 and 5.

### 3.1 Training

First, dataset is grouped by the nominal attribute with largest gain ratio without concerning attribute dependency. However a gain ratio based decision tree can replace single attribute grouping if dependency relationship is known. Grouping helps analyst name the most significant nominal property in helping classification.

Second, the grouped datasets are normalized and fed to KM-NC sub-model. Grouped datasets is clustered by K-means clustering for second-order paralleling computing after grouping, and more detailed structural knowledge of grouped dataset, upon which usage statistical methods can be used. Centroids are stored for forecast.

Third, one strong nonlinear classifier (MLP, GPC) is built for each clustered dataset. Clustered dataset is normalized again to train the NC model.

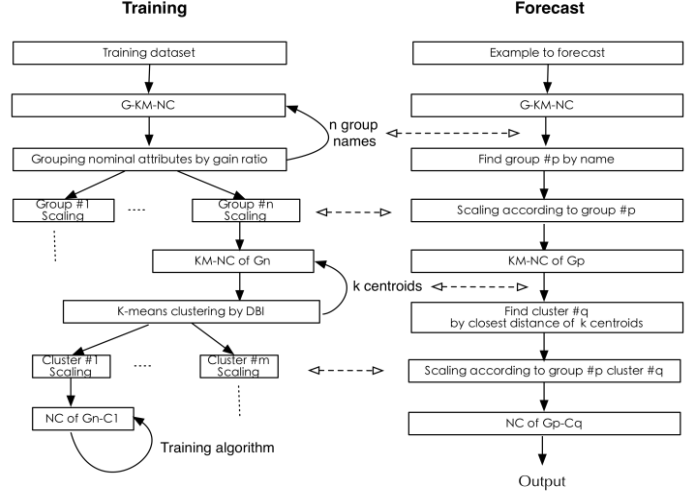

**Fig. 1.** G-KM-NC model

### 3.2 Forecast

First, the corresponding nominal group is found and data example is normalized according to preprocessing scheme used for training that group. Second, the closest cluster in the KM-NC is found by finding the closest centroid with data example.

$$D_j = \arg\min_{1 \leq j \leq k}\{\|v_{input} - c_j\|\} \qquad (2)$$

, where $v_{input}$ is the input, $c_j$ is the centroid of the $j^{th}$ cluster. Lastly, data example is normalized again according to the preprocessing scheme of the closest cluster and fed to non-linear classifier for outputting the result.

## 4 Criteria and Evaluation

In this section, we give the criteria of grouping and clustering process and evaluation.

### 4.1 Criteria for grouping and clustering

Information Gain is not normalized so the decision tree is prompt to have very wide subtree. In order to compensate for this, Quinlan suggests using the Gain Ratio as defined in [15]. Besides, a well-known internal criterion Davies-Bouldin Index (DBI) [22] is used to evaluate clustering. Smaller DBI value gives more significant clustering. We present that gain ratio and DBI well evaluate the grouping and clustering in our model in Section 5.

### 4.2 Result evaluation

As MLP produces a class output without any possibility, forecasted class is class 1 if output of MLP > 0.5 otherwise forecasted class is class -1. And besides, GPC produces possibility guess, and in binary classification situation, sum of possibility of both classes is 100%, we regard forecasted class is class 1 if output of GPC > 50%, and otherwise class -1. Accuracy of both nonlinear classifiers is:

$$Acc = \frac{n_{true-positive} + n_{true-negative}}{n} \quad (3)$$

Since G-KM-NC classifies an example by either MLP or GPC in the end, it can be treated as a new non-linear classifier using the same accuracy formula (3).

### 4.3 Model evaluation

In order to evaluate our model, we perform four main experiments. In the first one, four universal non-linear classifiers such as decision tree, Bayes network, MLP and Gaussian Process were carried out on different datasets. We will compare their performance with our model. Next, we evaluate the grouping part of our model (cf. Section 3, page 4). Performance of proposed model is shown in our third experiments. Finally, we also test the scalability of this model through its speedup performance.

## 5 Experiments and Analysis

Three well-known datasets belonging to different financial topics are tested in this paper. German credit dataset [23] uses 10 fold cross validation to form 900 training cases and 100 validation cases. Churn dataset [24] provides 3333 training and 1667 validation examples, in which phone number feature is dropped since it is unique by every customer and contains no useful info. House price dataset [25] uses 10 fold cross validation as well, consists of 2633 training cases and 293 validation cases.

MLP structure x-y-1 indicates it has x input neurons and y hidden layer neurons, in which optimal y is a priori. Both MLP and GPC use linear logistic function as activation function and interval/golden section search based conjugate gradient optimization method [19]. Step size and tolerance value of interval and golden section search are both 0.01.

### 5.1 Universal nonlinear classifier

We give validation performances of 4 universal nonlinear classifiers as baseline performances: C4.5, Bayes Network (BN), MLP and GPC. MLP picks the optimal hidden layer size from experiments to avoid both overfitting and under-fitting. GPC for churn dataset and house price dataset are omitted since the $O(n^3)$ complexity of GPC makes training impracticable. MLP and GPC use only numerical attributes of the datasets. From Table 1, models using more nominal labels (C4.5, Bayes Network) do not always outperform numerical models (MLP, GPC).

**Table 1.** Performance of universal nonlinear classifier

| Test Accuracy | German credit dataset | Churn dataset | House price dataset |
|---|---|---|---|
| C4.5 | 71.00% | 86.44% | 85.32% |
| Bayes Network | 73.00% | 88.00% | 82.90% |
| Optimal MLP | (7-3-1) 73.00% | (15-5-1) 93.34% | (7-5-1) 89.76% |
| GPC | 72.00% | / | / |

### 5.2 Grouping

We test the grouping by using only one nominal attribute. Structure so far is G-NC. We use the same hidden layer size among grouped datasets, assumption that grouped datasets by the same nominal attribute are of the same spatial distributive complexity.

**Table 2.** Grouping German credit dataset

| Att | Gain Ratio | Optimal G-MLP | G-GPC | Rank |
|---|---|---|---|---|
|  | ---------No-------- | (7-3-1) 73.00% | 72.00% |  |
| 1 | 0.05257(1st) | (7-3-1) 76.00% | 76.00% | Both 1st |
| 3 | 0.02548 | (7-3-1) 73.00% | 73.00% |  |
| 4 | 0.00934 | (7-2-1) 73.00% | 71.00% |  |
| 6 | 0.01666 | (7-1-1) 74.00% | 73.00% |  |
| 7 | 0.00608 | (7-1-1) 73.00% | 72.00% |  |
| 9 | 0.00445 | (7-2-1) 70.00% | 70.00% |  |
| 10 | 0.00891 | (7-1-1) 73.00% | 72.00% |  |
| 12 | 0.00872 | (7-3-1) 73.00% | 70.00% |  |
| 14 | 0.01051 | (7-1-1) 74.00% | 72.00% |  |
| 15 | 0.01120 | (7-1-1) 74.00% | 74.00% |  |
| 17 | 0.00095 | (7-5-1) 72.00% | 72.00% |  |
| 19 | 0.00099 | (7-1-1) 71.00% | 72.00% |  |
| 20 | 0.02550(2nd) | (7-1-1) 75.00% | 75.00% | Both 2nd |

From Table 2, we notice that quality of a grouping for understanding the behavior of financial dataset is related to the gain ratio of that nominal attribute. If GainRatio > 0.01, then forecast capability tends to improve. If GainRatio < 0.01, then grouping usually does not improve the forecast. A higher GainRatio indicates a better grouping model generally. 0.01 is a threshold for figuring if a grouping by a nominal attribute is significant or pointless for classification by MLP and GPC. Moreover, performance of GPC and optimal MLP are close and good G-GPCs outperform universal models.

Churn dataset and House price dataset are used to confirm grouping process. Since complexity of GPC is high, we test MLP only for 2 datasets. From Table 3 churn section, only grouping by the 5th attribute helps improve the predictive capability for churn dataset while grouping by 6th attribute keeps the same accuracy when the GainRatio ≫ 0.01. For the house price section, same phenomenon is listed. And G-MLPs by GainRatio>0.01 at least draw with the best universal models.

Indeed, grouping in the G-KM-NC model gives a first order parallel ability. If nominal attribute dependency is further known, it is reasonable to group the dataset by multilayer decision tree. The gain ratio distinguishes between important and non-critical features, which are an information measure feature selection according to [16]. The grouping not only outperforms universal models on validation accuracy but also disperses the computing pressure by either MLP or GPC.

**Table 3.** Grouping churn dataset and house price dataset

| churn dataset | | | house price dataset | | |
|---|---|---|---|---|---|
| Att | Gain Ratio | G-MLP | Att | Gain Ratio | G-MLP |
| ---------No-------- | | (15-5-1) 93.34% | ---------No-------- | | (15-5-1) 89.76% |
| 1 | 0.00318 | (15-5-1) 81.10% | 1 | 0.09022 | (15-3-1) 89.76% |
| 3 | 2.549E-5 | (15-3-1) 91.78% | 4 | 0.09709 (3$^{rd}$) | (15-3-1) 90.44% |
| 5 | 0.08032 (1$^{st}$) | (15-10-1) 94.90% | 5 | 0.16425 (2$^{nd}$) | (15-7-1) 90.44% |
| 6 | 0.00966 | (15-10-1) 93.34% | 6 | 0.05429 | (15-5-1) 89.76% |
| | | | 8 | 0.17029 (1$^{st}$) | (15-7-1) 90.44% |
| | | | 14 | 0.00074 | (15-3-1) 89.08% |
| | | | 15 | 0.00026 | (15-3-1) 89.76% |

### 5.3 Model performance

Next, we discover the inner distribution structure within the grouped dataset to gain more structural knowledge of the financial dataset. K-means clustering is used because it does not exclude much noise since recorded financial examples should be trusted generally. Furthermore, k-means clustering itself can be paralleled by [17].

In Table 4, A11, A12, A13 and A14 are the 4 nominal labels of No.1 attribute of German dataset, the numbers in brackets are the numbers of test cases of each group by cross validation. Lowest DBIs give the best accuracies over 4 groups, since lower DBI gives lower average similarity between clusters, indicating more significant clustering. However, it is not appropriate to set a universal threshold for clustering signif-icance because business varies. G-KM-NC discards gap areas between clusters in exchange of parallel performance improvement and extricates mutual interference of the classification surfaces of different clusters. G1-[7,8,5,5]-GPC gives the best accuracy thus far for German dataset:

$$\frac{26 \times 69.23\% + 43 \times 90.70\% + 27 \times 62.96\% + 4 \times 50\%}{26 + 43 + 27 + 4} = 76\% \quad (4)$$

Table 4. G1-KM-GPC of German credit dataset

| K | G1-KM-GPC for German credit dataset | | | | | | | |
|---|---|---|---|---|---|---|---|---|
|   | A12 (26) | DBI | A14 (43) | DBI | A11 (27) | DBI | A13 (4) | DBI |
| 1 | 69.23% |      | 90.70% |      | 59.26% |      | 50% |      |
| 2 | 65.38% | 1.76 | 90.70% | 1.96 | 62.96% | 1.91 | 50% | 1.92 |
| 3 | 69.23% | 1.66 | 90.70% | 1.34 | 48.15% | 1.59 | 50% | 1.80 |
| 4 | 61.54% | 1.58 | 88.37% | 1.49 | 51.85% | 1.76 | 50% | 1.37 |
| 5 | 69.23% | 1.52 | 88.37% | 1.42 | 62.96% | 1.46 | 50% | 1.16 |
| 6 | 61.54% | 1.47 | 88.37% | 1.42 | 59.26% | 1.61 | 50% | 1.34 |
| 7 | 69.23% | 1.22 | 86.06% | 1.35 | 59.26% | 1.60 | 50% | 1.58 |
| 8 | 65.38% | 1.24 | 90.70% | 1.30 | 48.15% | 1.53 | 25% | 1.52 |

Results of churn and house price dataset are listed in Table 5, in which International Plan(IP) is the 5[th] attribute of churn dataset and Central Air (CA) is the 8[th] attribute of house price dataset. Since datasets are large and GPC is impracticable, MLP is used as nonlinear classifier with the optimal hidden layer size in brackets and assuming clusters in one group have same distribution complexity. G5-[2,2]-MLP for churn dataset gets 95.62% overall accuracy and G8-[2,7]-MLP for house price dataset gets 90.78% accuracy. All three models outperform any basic universal model.

Table 5. G-KM-MLP of churn and house price datasets

| K | G5-KM-MLP churn dataset | | | | G8-KM-MLP house price dataset | | | |
|---|---|---|---|---|---|---|---|---|
|   | IP=no (1517) | DBI | IP=yes (150) | DBI | CA=Y (276) | DBI | CA=N (17) | DBI |
| 1 | (10)96.84% |      | (10)75.33% |      | (7)90.58% |      | (7)88.24% |      |
| 2 | (9)97.10%  | 1.85 | (10)80.67% | 1.88 | (4)90.94% | 1.09 | (7)88.24% | 1.60 |
| 3 | (9)96.77%  | 2.46 | (9)78.00%  | 2.32 | (4)90.22% | 1.24 | (7)88.24% | 1.51 |
| 4 | (7)95.85%  | 2.29 | (9)75.33%  | 2.31 | (3)90.58% | 1.23 | (7)88.24% | 1.62 |
| 5 | (7)95.19%  | 2.66 | (9)72.00%  | 2.29 | (3)90.22% | 1,32 | (3)88.24% | 1.68 |
| 6 | (7)96.18%  | 2.55 | (9)66.67%  | 2.48 | (3)90.22% | 1.45 | (2)88.24% | 1.48 |
| 7 | (7)95.39%  | 2.45 | (9)69.33%  | 2.35 | (3)90.22% | 1.38 | (1)88.24% | 1.45 |
| 8 | (5)95.72%  | 2.40 | (9)63.33%  | 2.23 | (3)89.86% | 1.38 | (1)88.24% | 1.45 |

Grouped dataset clustering gives a second order parallel ability by dispersing computing pressure further. It gives more detailed structure about groups. Clustering im-proves the validation accuracy when clustered by lowest DBI. If lowest DBI is not obvious, it indicates the group is tight and cannot be partitioned further like CA=N group in house price. If lowest DBI obviously exists, clustering by other K of higher DBI is illogical and will reduce the predictive capability. It is suggested that when the group is very large and optimal K obviously exists, use clustering, otherwise, skip it.

### 5.4 Speedup Analysis

GPC has high complexity $O(n^3)$ dominated by Cholesky decomposition when most computations are at most $O(n^2)$ [20], where n is the number of training data samples. It becomes impractical to train universal GPC when the dataset is large. If grouping or clustering partitions a dataset into p sub-datasets, the expected GPC complexity per thread is $O((n/p)^3)$, which lowers the overall complexity by $p^3$ times with p parallel threads. G-KM-MLP, in the same way, reduces BP training complexity from $O(n^2)$ [18] to $O((n/p)^2)$ per thread. Experiments use computers of 2.7 GHz Intel Core i5, 4GB 1333MHz DDR3 memory, and 4 cores. Results in Table 6 show our model is scalable and greatly improves the performance for both GPC and MLP as nonlinear classifier with multi-threading paradigm.

**Table 6.** Speedup Table

| Dataset | Model | No. Threads | Average time per thread |
|---|---|---|---|
| German | Universal GPC | 1 | 57764 seconds |
|  | G1-[7,8,5,5]-GPC | 25 | 71 seconds |
| Churn | Optimal Universal MLP | 1 | 1118 seconds |
|  | G5-[2,2]-MLP | 4 | 323 seconds |
| House price | Optimal Universal MLP | 1 | 509 seconds |
|  | G8-[2,7]-MLP | 9 | 10 seconds |

## 6  Conclusion and Future Work

In this paper, we present an integrated classification model, G-KM-NC, helping analyzing different financial datasets, referenced to practical data mart storage and the cognitive need of group/cluster structure. This model is a combined of different data mining techniques: grouping based on gain ratio, clustering and non-linear classification (MLP and GPC). Evidence that G-KM-NC outperforms other single-technique based universal model is presented and efforts are made to reduce the computing complexity by paralleling logically. Through our model, expert can not only understand financial dataset structurally, but also gain a good forecast capability. G-KM-NC model is flatter compared to DT, more fixed structure than BN whose structure is different between different fields, and more lightweight than universal MLP and GPC, more accurate than universal classifier techniques mentioned in this paper (DT, MLP, BN, GPC) and it uses only one single nominal attribute instead of all ones. G-KM-GPC outperforms G-KM-MLP by providing class possibility for a class forecast and does not need a priori knowledge. The main drawback of GPC is $O(n^3)$ complexity. We will explore the precise scope of G-KM-GPC model and introduce MLP stacked generalization to lower the computational burden of GPC or find a scalable GPC scheme.

Indeed, we also show that G-KM-NC gives a good parallel structure in classifying financial datasets. With its multi-threading approach, it is scalable for analyzing large datasets on high performance platforms [26].

Our designed model is well suitable for being used against star schema business data mart to tell which dimension tables are meaningful in predicting while others are not. However practical data mart size in business companies or banks may be over 100,000,000, which indicates G-KM-GPC method only suits small business environment. Our future work includes hierarchical grouping the larger dataset, which requires more dependency knowledge about a particular field, or further hierarchical clustering, which explores the inner sub-structure of clustered datasets.